\documentclass{article}

 \usepackage[preprint]{neurips_2026}

\usepackage[utf8]{inputenc} % allow utf-8 input
\usepackage[T1]{fontenc}    % use 8-bit T1 fonts
\usepackage{hyperref}       % hyperlinks
\usepackage{url}            % simple URL typesetting
\usepackage{booktabs}       % professional-quality tables
\usepackage{amsfonts}       % blackboard math symbols
\usepackage{nicefrac}       % compact symbols for 1/2, etc.
\usepackage{microtype}      % microtypography
\usepackage{xcolor}         % colors
\usepackage{amsmath}
\usepackage{multirow}
\usepackage{graphicx}
\usepackage{tikz}
\usepackage{amssymb}
\usepackage{xparse}
\usepackage{subcaption}
\usepackage{listings}
\usepackage{algorithm}

\title{Building Deep Graph Predictors with\\Graph Imitation Learning}

\author{%
    André Eberhard \quad Gerhard Neumann \quad Pascal Friederich \\
    Karlsruhe Institute of Technology (KIT) \\
    Karlsruhe, Germany \\
    \texttt{\{andre.eberhard,gerhard.neumann,pascal.friederich\}@kit.edu}
}

\begin{document}
    \maketitle

    \begin{abstract}
    Recent years have seen substantial progress in neural generation of text, images, and audio,
    supported by mature training pipelines and large-scale optimization.
    For graphs, however, comparable progress has been more limited. We attribute this gap
    to graph-specific optimization and representation challenges that undermine the effectiveness of training neural networks with backpropagation and gradient descent.
    We argue that representing graphs on a fixed-size Euclidean grid, as is common in recently proposed
    models for supervised graph prediction, may not be the optimal choice in these settings.
    To support our view, we provide an analysis of neural graph generation methods and identify theoretical challenges that
    lead to pitfalls when training neural networks to produce graphs as their output.
    Motivated by this analysis, we introduce \textbf{GRA}ph~\textbf{I}mitation~\textbf{L}earning~(GRAIL),
    a framework for training neural networks in supervised settings in which the supervision signal is
    a graph.
    GRAIL generates graphs sequentially through a Markov decision process over embeddings of partial graphs,
    thereby avoiding the representation issues associated with fixed-size grid graph representations.
    We empirically show that GRAIL achieves competitive results on supervised graph prediction across a
    comprehensive suite of 18 benchmarks, matching or surpassing state-of-the-art methods in several settings.
    \footnote{We make code, datasets and model checkpoints publicly available at \url{https://github.com/c72bcbf4/grail}}
\end{abstract}

    \section{Introduction}\label{sec:introduction}
We study the problem of supervised graph prediction.
The prediction of graphs from an input covers a wide variety of tasks and domains, such
as the generation of molecules~\cite{jtvae} or the recognition of their drawings
from images~\cite{clevert2021img2mol,qian2023molscribe,staker2019molecular,morin2023molgrapher},
the prediction of scene graphs from visual inputs~\cite{krishna2017visual,xu2017scene},
the recognition of road networks from aerial images~\cite{bastani2018roadtracer}, and autonomous
driving~\cite{buchner2023learning,zurn2023autograph,lv2025t2sg}, to name a few.
While models for text, images, and sound have improved dramatically in recent years,
driven by architectural improvements and large-scale optimization, comparable progress has been less pronounced for graph-structured data.
Unlike other modalities, learning to predict graphs introduces graph-specific issues
rooted in how graphs are represented.

The fundamental issues of supervised graph prediction are related to how neural networks represent graphs and how they are trained to predict graphs as their outputs.
Because of graph isomorphism, a graph with $n$ nodes can have up to $n!$ distinct but equivalent representations,
since its node order can be permuted without changing the represented graph.
With many equivalent representations, the optimization target becomes a whole isomorphism class rather than a single point in space.
As a result, the ambiguous supervision signal prevents direct end-to-end supervision of the predicted graph.
The community has made several attempts to address graph isomorphism for supervised training of graph outputs,
mainly under the framework of autoencoders for graphs~\cite{graphvae,winter2021pigvae} and supervised graph
prediction~\cite{any2graph2024,grale2025}.
These methods address graph isomorphism by fixing the alignment of nodes between the predicted and labeled graphs,
either to align their node order for point-wise supervision or to produce an already aligned output by learning the order internally.
They must do so to compensate for the architectural choice of representing graphs on a fixed size grid with feedforward networks,
inheriting the aforementioned issues.

In this work, we investigate an alternative formulation of the graph prediction problem.
We believe that treating graphs holistically and building permutation invariance into graph decoders is essential rather
than a design decision, and as such, a general and powerful method should not aim to use a factored representation for
graphs nor learn an internal ordering for an unordered modality.
In Section~\ref{sec:related-work} we provide an overview of popular methods for generating graphs, which
serves as a conceptual foundation of how different models address the generation of graphs.
In Section~\ref{sec:towards-permutation-invariant-decoding-of-graphs} we provide both theoretical and empirical
evidence that learning to align nodes for supervision may introduce issues in practice.
Following our analysis, in Section~\ref{sec:grail} we present our framework,
\textbf{GRA}ph~\textbf{I}mitation~\textbf{L}earning~(GRAIL), in which neural networks can be trained to
predict in various supervised settings.
We introduce a sequential, search-like generation procedure over partial graphs and propose a learning
framework for training neural networks to reconstruct a desired graph one element at a time.
By representing graphs as a unit quantity in the form of permutation-invariant graph embeddings, we can work around the optimization
issues graph isomorphism introduces.
To support our view, in Section~\ref{sec:results} we empirically evaluate models trained under our framework on common graph prediction benchmarks and beyond,
showing that they achieve competitive performance and match or surpass state-of-the-art graph-level decoders in several settings.
Finally, in Section~\ref{sec:limitations} we discuss the limitations of the proposed framework and outline directions for future research.

    \section{Related work}\label{sec:related-work}
In this paper, we study the problem of supervised graph prediction~\cite{krzakala2024framework}, a setting
in which the goal is to predict a target graph from an input.
This setting is distinct from graph generative modeling, in which the objective is to learn to generate
graphs similar to those of a reference distribution.
However, as both supervised and generative models need to derive procedures for generation,
the literature on generative models of graphs serves as the conceptual foundation of decoder
design, in particular regarding one-shot versus sequential construction and mechanisms for
handling graph representation and isomorphism as well as validity constraints.

\textbf{One-shot vs. sequential generation.}
One-shot methods represent the output graph as a single object in Euclidean space and optimize a graph's output representation with point-wise losses
for reconstruction, similar to images.
While these approaches are computationally efficient and simple to implement with standard feedforward architectures,
they inherently output an ordered representation of the graph and thus need to account for graph isomorphism during training.
As a consequence, one-shot methods typically use some form of graph matching or permutation-invariant
losses~\cite{graphvae,winter2021pigvae,any2graph2024,grale2025}.
Otherwise, optimizing decoders with point-wise losses for reconstruction would not be meaningful due to an ambiguous loss signal.
We provide further theoretical insights on the issues of point-wise reconstruction of graphs
in Section~\ref{sec:towards-permutation-invariant-decoding-of-graphs}.

Since graph isomorphism is a challenging problem to solve, another line of research focuses on generating graphs
sequentially~\cite{graphrnn,gran,learning-deep-gen-graphs,grover2019graphite,gcpn}.
By adding one or more elements at a time, these models learn to predict the next step recursively by taking into account
partial results.
While sequential inference avoids explicit graph matching during training, sequential generation inherently depends on a chosen construction
scheme and remains sensitive to how graphs are linearized.
As such, sequential generation procedures still need to account for the ordering implicitly by choosing which sequences of
operations to present to the model~\cite{graphrnn,learning-deep-gen-graphs}.
Furthermore, they must devise a generation procedure involving several forward passes for inference as well as step-wise supervision
of the process for training.
Their recursive nature makes generation inherently slower than one-shot generation but allows for more flexible and controlled generation.

Bridging one-shot generation and sequential generation, iterative refinement methods such as diffusion models generate graphs by repeatedly denoising an initial structure~\cite{gdss,digress},
thereby avoiding explicit node-by-node linearizations while still requiring multi-step inference.

\textbf{Unconditional vs. conditional generation.}
Besides algorithmic differences in how to generate a graph, there are also differences in the objective of graph generation.
Generative models for graphs primarily aim to model an underlying distribution with the ultimate goal
of being able to sample from this distribution, while they focus less on representing single samples exactly.
As such, we call this branch \emph{unconditional} generation.
This form of generation is mature and well-studied, and the earliest approaches date back to Erd\H{o}s--R\'enyi~\cite{erdos1960evolution} and
Barab\'asi--Albert~\cite{barabasi1999emergence} graphs.
Unfortunately, rule-based models are unable to fully capture the diversity of arbitrary graph classes, and as a result,
deep generative models of graphs have emerged.
In these models, neural networks provide the necessary flexibility to learn complex distributions of graphs in one-shot and sequential
architectures~\cite{vgae,gran,graphrnn,graphnvp,graphdf}.
Without further modification, these methods are unable to generate a specific graph given their non-deterministic nature due to
the ultimate goal of sampling diverse sets of graphs.

In contrast, methods for \emph{conditional} graph generation are also an important class of models as they may be used
in a wide spectrum of tasks in which the objective is to output a particular graph.
For example, models generating graphs conditioned on an image embedding can serve as image-to-graph translators, while
models generating graphs conditioned on a graph embedding can serve as graph-to-graph translators.
Therefore, another line of research focuses on such models.
Building on the seminal work of variational graph autoencoders~\cite{vgae}, which only reconstruct the graph
from node-level embeddings, the authors of~\cite{graphvae} propose the first graph-level extension.
They recognize that graph isomorphism is the inherent problem of reconstructing graphs in the autoencoder framework
and propose to introduce graph matching to find node correspondences and to calculate pointwise losses as in regular
autoencoders.
In~\cite{winter2021pigvae}, the authors introduce a permutation-invariant graph variational autoencoder (PIGVAE),
which goes one step further and establishes an architectural design in which the model learns a canonical ordering internally.
The authors argue that because during global aggregation, the node order is lost in the graph latent, a model needs to learn to infer it.
They address this issue by introducing an additional predictor of the graph's node count as well as a permuter model that
predicts the ordering of these nodes with a learned SoftSort operator~\cite{prillo2020softsort}.
However, as we explain in Section~\ref{sec:towards-permutation-invariant-decoding-of-graphs}, treating the node ordering
as a latent variable which can be inferred may be problematic when the ordering is not unique and inconsistent between samples.
More recently, a line of research has been proposed under the term of supervised graph prediction, in which models directly
translate inputs of various modalities into graphs~\cite{krzakala2024framework,any2graph2024, grale2025}.
In~\cite{any2graph2024}, the authors seek to develop a permutation-invariant distance for graphs and introduce a loss based
on optimal transport that is permutation-invariant and handles variable-sized inputs.
A notable improvement is that their model, Any2Graph, can predict graph-structured outputs not only from graphs but also from images
and fixed vectors.
In follow-up work~\cite{grale2025}, the authors combine the loss introduced for Any2Graph and the matching strategy of PIGVAE\@.
The resulting architecture, GRALE, operates on both graph and node embeddings, where the graph-level embeddings serve as a latent
representation and node-level embeddings are used for matching, achieving state-of-the-art results. Limitations of these approaches will be discussed in the next section.

Although contemporary distributional generators such as diffusion models~\cite{gdss,digress} or normalizing flows~\cite{graphnvp,graphdf}
can be adapted for conditional generation, their objective remains distribution matching.
Because they are trained to recover data distributions through sampling, they are generally not used for direct latent-to-graph
decoding where the goal is exact reconstruction of a specific target graph.

    \section{Analysis of existing methods for supervised graph prediction}\label{sec:towards-permutation-invariant-decoding-of-graphs}
In this section, we aim to provide the necessary intuition for desirable and undesirable components for general graph decoders.
Then, building on this intuition, we present our proposed framework, GRAIL, in more detail in Section~\ref{sec:grail}.

To address the representational ambiguities caused by graph isomorphism during optimization,
contemporary graph decoding models such as Any2Graph~\cite{any2graph2024} and GRALE~\cite{grale2025} learn an ordering internally.
However, they optimize the output w.r.t.\ a single ordering $\pi$ of a whole isomorphism class $\Pi$.
While optimizing w.r.t. $\pi$ works empirically, the implications of using a single $\pi_{n} \in \Pi$ as target
remain unclear, as any other $\pi \in \Pi \setminus \pi_{n}$ would be an equally valid target.
With an increasing number of samples, it becomes more and more difficult for the model to commit to a particular ordering
internally as the underlying, unobserved $\pi$ is likely to be different for different samples, which creates the following conflicts.
For example, given two embeddings of two non-isomorphic graphs of the same size, their target representation may be ordered by
different $\pi$.
During training, the network must learn to associate pairs $(G_{1},\pi_{1})$ and $(G_{2},\pi_{2})$ whereas $\pi_{1} \neq \pi_{2}$,
which creates an internal conflict.
However, as $\pi$ is not represented and ambiguous there is no shared structure to learn since a canonical ordering is missing.
Under such circumstances, due to a missing consistency in the labeling process, it is more likely that the network learns instance-specific
solutions rather than generalized ones, as neural networks are known to be able to fit data well even if there are no consistent
labels~\cite{zhang2016understanding}.
To overcome this limitation, it may be tempting to optimize w.r.t.\ (a subset of) $\Pi$ as a target, e.g., by optimizing w.r.t.\
the restricted class of breadth-first search (BFS) orderings as in GraphRNN~\cite{graphrnn}.
However, given a latent representation $\mathbf{z}$, optimizing w.r.t.\ multiple $\pi_{n} \in \Pi$ also creates a conflict
as the ordering is not represented in $\mathbf{z}$ but multiple targets exist.
For example, when learning with point-wise losses in such a scenario, a target ordering $\pi_{1}$ may push the
network to predict the existence of a node for a particular slot, whereas a target ordering $\pi_{2}$ may push the network to assign no
node to the same slot.

To validate our claims empirically, we conduct the following toy experiments.
For three datasets of small graphs, we train a graph neural network (GNN) encoder and a multi-layer perceptron (MLP)
decoder end-to-end to predict the adjacency matrix for the graphs.
While the graphs come with both node and edge features, we only train to predict the adjacency
matrix for illustration purposes and also fix the graph size to nine nodes such that the output dimension remains fixed
and to make it comparable between all datasets.
The graphs then undergo different transformations of their ordering, namely random order, BFS ordering from the highest degree node, and canonical
ordering as implemented in the igraph~\cite{csardi2006igraph} library.
We train on 1M samples and report the overall loss and the final reconstruction accuracy on 10,000 samples.
A description of the used dataset can be found in Section~\ref{sec:results}.
The results of our experiments can be found in Table~\ref{tab:ordering_comparison}.

\begin{table}[h]
    \centering
    \begin{tabular}{|c|ccc|ccc|}
        \hline
        \multirow{2}{*}{Dataset} & \multicolumn{3}{c|}{Crossentropy loss} & \multicolumn{3}{c|}{Reconstruction accuracy} \\
        \cline{2-7}
        & Random & BFS    & Canonical & Random & BFS    & Canonical \\
        \hline
        TREES    & 0.4710 & 0.4711 & 0.0093    & 0.0    & 0.0    & 0.9550    \\
        COLORING & 0.6150 & 0.6150 & 0.0597    & 0.0    & 0.0    & 0.7698    \\
        QM9      & 0.1286 & 0.1218 & 0.1221    & 0.1920 & 0.1872 & 0.3267    \\
        \hline
    \end{tabular}
    \caption{Comparison of ordering strategies across datasets for loss and accuracy.}
    \label{tab:ordering_comparison}
\end{table}

Our argument is supported in two meaningful ways.
First, there is a notable difference in performance between random or heuristic orderings
and a canonical ordering.
This result is not surprising as by our argument, a consistent ordering provides a consistent
update direction and allows the model to detect patterns in the data.
Similarly, simple heuristics like BFS orderings based on global node statistics are not better than
random as simple heuristics still fail to impose a meaningful constraint on the node order.
Second, the reconstruction accuracies for different graph classes show notable differences after
training for 1M samples.
While trees, the simplest class of graphs in this experiment, achieve good reconstruction
fidelity, the models fail to do so on moderately more complex instances.

In summary, we postulate that assigning a single, random permutation $\pi$ to each sample induces inter-sample conflicts
as the underlying ordering of different samples yields inconsistent update directions, while using several
$\pi$ for each sample induces intra-samples conflicts.
Crucially, models such as Any2Graph~\cite{any2graph2024} and GRALE~\cite{grale2025} are trained with losses
which match against a single, non-canonical $\pi$, making the loss calculation permutation invariant.
However, their decoder components are not permutation invariant as they inherently output an ordered representation
which is trained against a specific $\pi \in \Pi$ to reproduce this order.
This may cause the model to randomly fit a chosen ordering to a particular sample, which inherently limits the
models' generalization capabilities as there is no shared structure to learn from.
Given our theoretical and empirical analyses, we believe that exploring alternatives to ordering-based techniques
for supervised graph prediction may be worthwhile.

    \section{GRAIL: A framework for supervised graph prediction}\label{sec:grail}
In this section, we introduce our proposed framework, \textbf{GRA}ph \textbf{I}mitation \textbf{L}earning (GRAIL)
in four steps.
First, based on our analyses in Section~\ref{sec:towards-permutation-invariant-decoding-of-graphs},
we define desirable properties for a general graph decoding algorithm.
Second, we formalize the setup in which we learn neural network decoders for graphs.
Third, we discuss how to instantiate the setup, and fourth, how to train within our framework in practice.
We provide an extended discussion of the limitations and possible future improvements in Section~\ref{sec:limitations}.

\textbf{Desiderata for permutation-invariant graph decoding.}
Following our rationale in Section~\ref{sec:towards-permutation-invariant-decoding-of-graphs},
a general decoder should, first and foremost, operate fully at the graph level in order to avoid
the issue of graph isomorphism, which comes along with several difficult theoretical issues that are
hard to resolve in practice.
As such, we aim for solutions that do not rely on modeling the graph's nodes and edges but rather
treat graphs holistically as one unit and in a permutation-invariant way.
Second, the method should support variable-sized graphs naturally as datasets of graphs usually
contain graphs of a diverse range of sizes.
Third, while not strictly required, it would be good to have validity guarantees
for graph generation, as models based on predicting single elements of a graph are brittle due to
the non-Euclidean nature of graphs.

To satisfy the first two conditions, we postulate that such a method should generate the graph
sequentially as this is an effective way to avoid the issue of graph isomorphism.
Furthermore, to treat the graph holistically and to naturally support variable-sized inputs,
such a method should operate solely at the graph embedding level to avoid the need to deal with
nodes and their orderings, as this is not the primary quantity we care about.
By operating fully at the level of graph embeddings we can avoid dealing with the choice
of graph representation, i.e., whether we use an adjacency matrix or tensor, adjacency list, incidence matrix
or edge list, and as such, the graph embedding becomes the single source of truth for its representation.
Therefore, we propose a sequential method that generates a graph one element at a time by predicting
fragments of graphs in a search-like procedure.
By modifying a partial result and predicting over its graph-level representation whether it lies on
the path towards the desirable solution, we are able to build the target graph one step at a time
without having to deal with the aforementioned issues.

\textbf{GRAIL as a Markov decision process.}
To decode sequentially at the graph-level, we train a model which, given an initially empty
graph $G_{0}$ and a target graph $G_{T}$, yields a trajectory of graphs
$(G_{0} \rightarrow G_{1}\rightarrow G_{t} \rightarrow \ldots \rightarrow G_{T})$.
To achieve this, we must define how to express the transition $G_{t} \rightarrow G_{t+1}$
and how to learn which of these transitions are valid.
We formalize our transition model within a Markov decision process (MDP) and optimize an
imitation learning (IL) objective as follows.
An MDP is typically defined as a 5-tuple \(\mathcal{M} = (\mathcal{S}, \mathcal{A}, P, R, \gamma)\),
where \(\mathcal{S}\) is the state space, \(\mathcal{A}\) the action space, \(P(s' \mid s,a)\) the
transition model, \(R(s,a)\) the reward function, and \(\gamma \in [0,1)\) the discount factor.
As we are in an episodic setting, we strictly use $\gamma=1$.
Furthermore, we define this MDP such that rewards $r \in \{0,1\}$ are only emitted at terminal
states indicating whether the correct graph has been generated.
The action space is induced by the sets $\mathbb{N}$ and $\mathbb{E}$ defining the available node and
edge types, respectively.
The action space $\mathcal{A} = (\mathbb{N} \times \mathbb{E} \times \mathbb{N}) \cup \{\tau\} $ consists of
all possibilities to connect two (typed) nodes with a (typed) edge as well as a special no-op
action $\tau$ indicating termination.
For example, in a colored graph domain where $\mathbb{N}=\{red,blue\}$ and $\mathbb{E}=\{-\}$,
the action triplet $(red,-,blue)$ indicates that a blue node may be attached to a red node via a
connection type $-$.
Given a graph $G_{t}$, we can then apply each possible action to each node in the graph, creating a set
$\mathbb{M}_{t}$ of successor states which differ by exactly one additional connection, representing
either a connection between two existing nodes or the addition of a new node to the graph.
We then express a state transition $G_{t} \rightarrow G_{t+1}$ by choosing one $G_{t+1} \in \mathbb{M}_{t}$ as the basis for
the next time step and iterate this process until a terminal action has been selected.
Note that in this MDP, we require actions to only add to and never delete edges from a graph,
which simplifies the learning algorithm we derive next.

\paragraph{Training a policy to predict valid transitions.}
We now define the validity of a transition $G_t \rightarrow G_{t+1}$ and describe how to
train neural networks to predict sequences of valid transitions under this definition.
For the remainder of this section, we refer to graphs $G_{t}$ as \emph{queries} and graphs $G_{T}$ as \emph{targets}.
We base our learning algorithm on the following insight.
It holds for any sequence of graphs $(G_{0} \rightarrow G_{1}\rightarrow G_{t}\rightarrow \ldots \rightarrow G_{T})$ terminating with an
exact match $G_{T}$ that $\forall t: G_{t} \subseteq G_{T}$, i.e., a correct trajectory
must always be a sequence of subgraphs, as any non-subgraph cannot be a valid partial result
if our procedure only adds elements at each time step.
Intuitively, if a target graph $G_{T}$ contains only three red nodes, adding both a fourth red node
or a node of any other color creates a successor state $G_{t+1}$ from which the policy
cannot recover when only additive actions are permitted.

Remember that in the supervised graph prediction problem, we aim to predict from an input
$x_{T}$ the associated target graph $G_{T}$.
Note that $x_{T}$ can be an arbitrary input such as an image, graph, or any other modality
for which one would like to predict the corresponding graph output.
To do this, we define three neural networks.
First, given an input $x_{T}$, we define a target encoder $f_{T}$ to produce a latent embedding $\mathbf{z_{T}}$.
Second, we define a query encoder $f_{Q}$ to produce a latent embedding $\mathbf{z_{Q}}$ of $G_t$.
Third, we define a policy as $\pi(\mathbf{z_Q}, \mathbf{z_T}, term) \rightarrow \{0,1\} $ where $term \in \{0,1\} $ is a binary flag indicating whether
an encoded query corresponds to a self-transition signaling termination.
Note that we defined our decision function to operate on latent representations for both query and target.
As a consequence, our formulation becomes task-agnostic w.r.t.\ the modality of inputs $x_T$.
For example, given a dataset of molecular graphs, $x_{T}$ could be the result of generating a molecular drawing
from the corresponding $G_{T}$ for image-to-graph tasks, or $x_{T}$ could equal $G_{T}$ for graph-to-graph tasks.
This distinction is important because during inference, the target graph $G_{T}$ is not available since
the prediction task aims to recover it from its representation $x_{T}$.

Next, we need to define how to train our policy to choose valid transitions.
We choose to train our models in a setup similar to DAgger~\cite{dagger}, which we do as follows.
Given our MDP and a dataset $\mathcal{D}$, we can sample a target $G_{T} \sim \mathcal{D}$ and obtain a representation
$x_{T}$, e.g., by rendering the graph if we would like to predict a graph from an image.
As we defined the validity of a transition from $G_{t} \rightarrow G_{t+1}$ to satisfy the subgraph property,
in this work, we use VF2++~\cite{juttner2018vf2++}, a widely used algorithm for subgraph isomorphism,
as the expert policy $\pi_{E}$.
We can then roll out the policy $\pi$ in the MDP we defined, and along the trajectories label
occurring successor states in $\mathbb{M}_{t}$.
We always use a greedy policy during both training and inference.
While our framework does not prescribe a particular training architecture, in this paper, we choose
to instantiate many trajectories in parallel and to learn the policy fully online.
In this training architecture, the policy is continuously improved with more rollouts, as each interaction
provides ground-truth data directly at the time of expanding states.

\paragraph{Filtering invalid and redundant actions.}
We would like to conclude this section by addressing several practical issues which are mainly related
to the data quality and thus efficiency of training.
In practice, several issues arise when naively training under the setup we described so far.
The first issue is related to the exploding search space at each step.
Given the set of node types $\mathbb{N}$ and edge types $\mathbb{E}$, together they induce an
action space $\mathcal{A}=\mathbb{N} \times \mathbb{E} \times \mathbb{N}$.
The action space describes which types of nodes may be connected with which types of edges, i.e., a
triple $(n_{u},e,n_{v})$ connects a node $u$ of a type $n_{u}$ to a node $v$ of a type $n_{v}$ via an edge of a type $e$.

However, especially in larger action spaces, creating $\mathbb{M}_{t}$ based on the full $\mathcal{A}$
is wasteful since not all graphs always contain all types of connection triplets.
For example, given a graph consisting only of one node type $n \in \mathbb{N}$ and one edge type $e \in \mathbb{E}$,
expanding actions other than $(n,e,n)$ certainly will not produce valid transitions, and we thus should not expand
these at all.
As a solution, we learn, in addition to the policy $\pi$, a filter network $f_{E}:(\mathbf{z}_{Q},\mathbf{z}_{T}) \rightarrow \{0,1\}^{|\mathcal{A}|}$
predicting for each possible action triplet $(n,e,n)$ in $\mathcal{A}$ whether we should perform the according modification
to create a successor state.
Note that this definition does not rely on the notion of a terminal state.
For example, if the target graph contains five connections in which a black edge connects two red nodes,
and the query contains the same number of the same connection type, we do not connect any more pairs of red
nodes with black edges.
This labeling process is trivial as we only need to count the number of occurrences for all triplets $(n,e,n)$ in both the query and
the target, calculate the difference and assign a binary label indicating whether we can still add more connections of this type.
Note by treating every $G_{t}$ along the trajectory like this, this not only filters actions which are generally
not possible but also those which have been satisfied so far, further reducing the search space.
We found this filter to be trivially learned in comparison to the IL objective, as the network must only
learn to associate distributional properties of the embeddings rather than structural ones.
Additionally, during inference, whenever an expansion $(n_{u},e,n_{v})$ was predicted as an invalid
transition, we do not expand it again in later steps.
Filtering the available actions before expansions and avoiding enumerating invalid states
repeatedly, we can keep the search space manageable.
We provide pseudocode for the complete procedure in Algorithm~\ref{alg:grail}.

\paragraph{Practical considerations for training.}
Another practical issue is related to the optimization process itself.
While the extent depends on the distributional properties of a particular dataset, the search-like nature of our
method cannot avoid training under imbalance and easy negatives.
Inherently, the number of all possible non-subgraphs is simply much larger and may distort the optimization process
under extreme imbalance.
While the action filter on $\mathcal{A}$ already removes the occurrence of the most trivial negative
examples, there still remains a large fraction representing easy negatives.
Whenever these conditions become too extreme, one may consider switching from binary cross-entropy to focal loss~\cite{lin2017focal}.
Focal loss is an effective technique in these scenarios as it suppresses the gradient signal from the easy majority
and keeps optimization in a stable regime.
Even though not necessary for all datasets, in the experiments in Section~\ref{sec:results}, we train all models with focal loss with $\gamma=3.0$
and without class-weights to present results with consistent hyperparameters across runs.
We train both the subgraph policy $\pi$ and filter network $f_{E}$ with a combined, unweighted focal loss.
Both predictors operate on the shared tuple $(\mathbf{z}_{Q},\mathbf{z}_{T})$ of latent representations obtained by $f_{Q}$ and $f_{T}$.

    \section{Results}\label{sec:results}

\textbf{Evaluating graph prediction accuracy.}
In this section, we evaluate our approach on a wide range of graph prediction tasks.
We follow the experimental setup of~\cite{grale2025} with the following modifications.
We generate samples for TREES, COLORING 15 and COLORING 20 in a streaming fashion
without a fixed dataset, which slightly differs from the setup in~\cite{grale2025}, which used a fixed
dataset of 300k samples, whereas in our setup this dataset is effectively much larger and therefore
more challenging to generalize.

We report the prediction accuracy over 10,000 samples in Table~\ref{tab:results}.
Across all datasets and modalities, our models demonstrate performance comparable to the state of the art and, in several
settings, surpass the best available baseline by a large margin.
For short trajectories, our models achieve near-perfect results, while larger and structurally more complex graphs are,
unsurprisingly, harder to predict.
Note that for generating fingerprints for synthetic graphs, we use the algorithm in~\ref{sec:experimental-setup},
which does not come with any guarantees of how accurately it encodes a particular graph into a high-dimensional, binary
vector, and thus may set an upper limit on decoding performance.
Furthermore, to make experiments comparable, we performed all runs with an identical configuration where
possible and with a fixed time-budget~(for more details see Appendix ~\ref{sec:appendix-training}).
On the COLORING datasets, our models show a lower performance level which we attribute to not tuning models per
experiment, time-bound training as well as an increased data volume potentially associated with bigger generalization challenges, rather than a small, fixed-size dataset.

In addition to prediction performance, we report timing and search statistics, namely the time required
to decode a full trajectory, the average number of successor states per transition, and the average predicted
and true trajectory lengths.
Even though our model generates and evaluates many graph pairs for each trajectory, the time to decode even the largest samples
is on the order of milliseconds in a batched setting.
While an affordable expert policy is important for generating training data, the search space is primarily a concern
for inference latency.
For all datasets, evaluating the expert policy takes no more than one millisecond for more than 99\% of the samples.
Highly regular, synthetic graphs such as those in the COLORING dataset are the most expensive ones to evaluate,
while more constrained real-world graphs show much slower growth in evaluation time.
Furthermore, we investigate the number of successor states per step and dataset.
Our proposed mitigations for the exploding search space effectively mitigate the exponential growth compared to
not filtering at all.
Across all datasets, the search space grows within roughly the first half of the trajectory.
In the second half, however, the search space tends to shrink rather than to grow further with length, showing the
effectiveness of the proposed filtering techniques (details in Appendix~\ref{sec:appendix-training}).
For much larger graphs, inference becomes the bottleneck in the current setup, and we provide more information on how we can optimize
inference further for large instances in Section~\ref{sec:limitations}.

\definecolor{taskgraph}{HTML}{188bb5}
\definecolor{taskimage}{HTML}{fc5603}
\definecolor{taskfp}{HTML}{1fa35a}

\NewDocumentCommand{\gcell}{s m}{%
    \textcolor{taskgraph}{\IfBooleanTF{#1}{\textbf{#2}}{#2}}%
}
\NewDocumentCommand{\icell}{s m}{%
    \textcolor{taskimage}{\IfBooleanTF{#1}{\textbf{#2}}{#2}}%
}
\NewDocumentCommand{\fcell}{s m}{%
    \textcolor{taskfp}{\IfBooleanTF{#1}{\textbf{#2}}{#2}}%
}

\begin{table}[htb]
    \scriptsize
    \centering
    \setlength{\tabcolsep}{2.5pt}
    \begin{tabular}{|c||c||c|c||c|c|c|c|c|}
        \hline
        Dataset & Task & Any2Graph & GRALE & GRAIL (ours) & $\varnothing$ Duration/Sample (s) & $\varnothing$ Successors & $\varnothing$ Steps pred. & $\varnothing$ Steps true \\
        \hline
        \textbf{TREES}
        &
        \begin{tabular}[c]{@{}c@{}}
            \gcell{graph} \\
            \icell{image} \\
            \fcell{fingerprint}
        \end{tabular}
        &
        \begin{tabular}[c]{@{}c@{}}
            \gcell{--} \\
            \icell{--} \\
            \fcell{--}
        \end{tabular}
        &
        \begin{tabular}[c]{@{}c@{}}
            \gcell{--} \\
            \icell{--} \\
            \fcell{--}
        \end{tabular}
        &
        \begin{tabular}[c]{@{}c@{}}
            \gcell*{99.79\%} \\
            \icell*{98.39\%} \\
            \fcell*{61.74\%}
        \end{tabular}
        &
        \begin{tabular}[c]{@{}c@{}}
            \gcell{0.0082} \\
            \icell{0.0247} \\
            \fcell{0.0060}
        \end{tabular}
        &
        \begin{tabular}[c]{@{}c@{}}
            \gcell{9.03} \\
            \icell{9.04} \\
            \fcell{13.28}
        \end{tabular}
        &
        \begin{tabular}[c]{@{}c@{}}
            \gcell{9.93} \\
            \icell{9.93} \\
            \fcell{7.95}
        \end{tabular}
        & \begin{tabular}[c]{@{}c@{}}
              \gcell{9.94} \\
              \icell{10.0} \\
              \fcell{9.86}
        \end{tabular}
        \\ \hline
        \textbf{COLORING 15}
        &
        \begin{tabular}[c]{@{}c@{}}
            \gcell{graph} \\
            \icell{image} \\
            \fcell{fingerprint}
        \end{tabular}
        &
        \begin{tabular}[c]{@{}c@{}}
            \gcell{--}      \\
            \icell{43.77\%} \\
            \fcell{--}
        \end{tabular}
        &
        \begin{tabular}[c]{@{}c@{}}
            \gcell{--}       \\
            \icell*{88.87\%} \\
            \fcell{--}
        \end{tabular}
        &
        \begin{tabular}[c]{@{}c@{}}
            \gcell*{93.78\%} \\
            \icell{70.80\%}  \\
            \fcell*{69.99\%}
        \end{tabular}
        &
        \begin{tabular}[c]{@{}c@{}}
            \gcell{0.0198} \\
            \icell{0.0330} \\
            \fcell{0.0147}
        \end{tabular}
        &
        \begin{tabular}[c]{@{}c@{}}
            \gcell{21.38} \\
            \icell{22.65} \\
            \fcell{23.18}
        \end{tabular}
        &
        \begin{tabular}[c]{@{}c@{}}
            \gcell{19.89} \\
            \icell{18.31} \\
            \fcell{17.62}
        \end{tabular}
        &
        \begin{tabular}[c]{@{}c@{}}
            \gcell{20.71} \\
            \icell{21.53} \\
            \fcell{21.17}
        \end{tabular}

        \\ \hline

        \textbf{COLORING 20}
        &
        \begin{tabular}[c]{@{}c@{}}
            \gcell{graph} \\
            \icell{image} \\
            \fcell{fingerprint}
        \end{tabular}
        &
        \begin{tabular}[c]{@{}c@{}}
            \gcell{--} \\
            \icell{--} \\
            \fcell{--}
        \end{tabular}
        &
        \begin{tabular}[c]{@{}c@{}}
            \gcell*{99.20\%} \\
            \icell{--}       \\
            \fcell{--}
        \end{tabular}
        &
        \begin{tabular}[c]{@{}c@{}}
            \gcell{86.72\%}  \\
            \icell*{57.76\%} \\
            \fcell*{58.89\%}
        \end{tabular}
        &
        \begin{tabular}[c]{@{}c@{}}
            \gcell{0.0280} \\
            \icell{0.0386} \\
            \fcell{0.0202}
        \end{tabular}
        &
        \begin{tabular}[c]{@{}c@{}}
            \gcell{29.96} \\
            \icell{32.23} \\
            \fcell{31.82}
        \end{tabular}
        &
        \begin{tabular}[c]{@{}c@{}}
            \gcell{23.92} \\
            \icell{19.72} \\
            \fcell{19.44}
        \end{tabular}
        &
        \begin{tabular}[c]{@{}c@{}}
            \gcell{26.46} \\
            \icell{26.91} \\
            \fcell{25.60}
        \end{tabular}
        \\ \hline

        \textbf{QM9}
        &
        \begin{tabular}[c]{@{}c@{}}
            \gcell{graph} \\
            \icell{image} \\
            \fcell{fingerprint}
        \end{tabular}
        &
        \begin{tabular}[c]{@{}c@{}}
            \gcell{--} \\
            \icell{--} \\
            \fcell{29.85\%}
        \end{tabular}
        &
        \begin{tabular}[c]{@{}c@{}}
            \gcell{--} \\
            \icell{--} \\
            \fcell{35.27\%}
        \end{tabular}
        &
        \begin{tabular}[c]{@{}c@{}}
            \gcell*{100\%} \\
            \icell*{100\%} \\
            \fcell*{100\%}
        \end{tabular}
        &
        \begin{tabular}[c]{@{}c@{}}
            \gcell{0.0084} \\
            \icell{0.0215} \\
            \fcell{0.0070}
        \end{tabular}
        &
        \begin{tabular}[c]{@{}c@{}}
            \gcell{12.28} \\
            \icell{11.55} \\
            \fcell{11.90}
        \end{tabular}
        &
        \begin{tabular}[c]{@{}c@{}}
            \gcell{10.43} \\
            \icell{10.45} \\
            \fcell{10.39}
        \end{tabular}
        &
        \begin{tabular}[c]{@{}c@{}}
            \gcell{10.43} \\
            \icell{10.45} \\
            \fcell{10.39}
        \end{tabular}
        \\ \hline

        \textbf{PUBCHEM 32}
        &
        \begin{tabular}[c]{@{}c@{}}
            \gcell{graph} \\
            \icell{image} \\
            \fcell{fingerprint}
        \end{tabular}
        &
        \begin{tabular}[c]{@{}c@{}}
            \gcell{--} \\
            \icell{--} \\
            \fcell{--}
        \end{tabular}
        &
        \begin{tabular}[c]{@{}c@{}}
            \gcell{66.80\%} \\
            \icell{--}      \\
            \fcell{--}
        \end{tabular}
        &
        \begin{tabular}[c]{@{}c@{}}
            \gcell*{90.66\%} \\
            \icell*{41.84\%} \\
            \fcell*{29.04\%}
        \end{tabular}
        &
        \begin{tabular}[c]{@{}c@{}}
            \gcell{0.0389} \\
            \icell{0.0510} \\
            \fcell{0.0341}
        \end{tabular}
        &
        \begin{tabular}[c]{@{}c@{}}
            \gcell{44.79} \\
            \icell{45.09} \\
            \fcell{61.99}
        \end{tabular}
        &
        \begin{tabular}[c]{@{}c@{}}
            \gcell{23.17} \\
            \icell{17.57} \\
            \fcell{15.70}
        \end{tabular}
        &
        \begin{tabular}[c]{@{}c@{}}
            \gcell{24.38} \\
            \icell{24.98} \\
            \fcell{24.94}
        \end{tabular}
        \\ \hline

        \textbf{PUBCHEM 40}
        &
        \begin{tabular}[c]{@{}c@{}}
            \gcell{graph} \\
            \icell{image} \\
            \fcell{fingerprint}
        \end{tabular}
        &
        \begin{tabular}[c]{@{}c@{}}
            \gcell{--} \\
            \icell{--} \\
            \fcell{--}
        \end{tabular}
        &
        \begin{tabular}[c]{@{}c@{}}
            \gcell{--} \\
            \icell{--} \\
            \fcell{--}
        \end{tabular}
        &
        \begin{tabular}[c]{@{}c@{}}
            \gcell*{63.26\%} \\
            \icell*{05.09\%} \\
            \fcell*{14.08\%}
        \end{tabular}
        &
        \begin{tabular}[c]{@{}c@{}}
            \gcell{0.07975} \\
            \icell{0.0649}  \\
            \fcell{0.0544}
        \end{tabular}
        &
        \begin{tabular}[c]{@{}c@{}}
            \gcell{71.75} \\
            \icell{64.17} \\
            \fcell{80.06}
        \end{tabular}
        &
        \begin{tabular}[c]{@{}c@{}}
            \gcell{30.28} \\
            \icell{17.36} \\
            \fcell{16.82}
        \end{tabular}
        &
        \begin{tabular}[c]{@{}c@{}}
            \gcell{37.49} \\
            \icell{37.58} \\
            \fcell{37.71}
        \end{tabular}
        \\ \hline

    \end{tabular}

    \caption{
        Accuracy of decoding \textcolor{taskgraph}{graphs}, \textcolor{taskimage}{images} and \textcolor{taskfp}{fingerprints} into
        graphs on various datasets. For Any2Graph~\cite{any2graph2024} and GRALE~\cite{grale2025}, the missing values have not been
        reported and no model checkpoints were release in the public repositories.
    }
    \label{tab:results}
\end{table}

\textbf{Evaluating embedding-level arithmetic.}
As shown in the previous paragraph, a graph-level decoder can serve as an elegant model for translating
various kinds of modalities into the underlying graph representation.
The previous experiments were conducted in an encoder-decoder setup where the goal was to
generate a particular graph from an input.
In this paragraph, we study the decoder-only properties of GRAIL in the context of graph edits
in the embedding space.
Being able to decode from an arbitrary embedding into an (approximately) correct version of the
embedded graph can serve various graph-level tasks, such as graph interpolation, modification,
or generation.

In particular, we are interested in whether our model produces a meaningful latent space, i.e.,
whether modifications of the latent embedding $\mathbf{z}$ produce meaningful variations in the
graph space.
Since we can decode graphs from arbitrary embeddings, we perform the following experiment.
We sample a graph and use the trained encoder to obtain a graph embedding $\mathbf{z}$ which
we use as a baseline for modification.
Starting from $\mathbf{z}$, we linearly interpolate between the origin $\mathbf{0}$ and $c\mathbf{z}$ for some $c>1$.
Decoding this set of embeddings allows us to inspect the resulting graphs visually.
We show several decoded examples in Figure~\ref{fig:interpolations}, where the top, middle, and bottom
rows represent samples from the COLORING~15, TREES, and PUBCHEM~32 datasets, respectively.

Our examples show that a larger embedding norm also corresponds to larger decoded graphs,
which indicates that all models encode a notion of size in the embedding space.
What is more interesting is the local neighborhood consistency.
In all cases, and across all embedding scales, the type of connections is fully consistent with
the baseline $\mathbf{z}$ for all scaled versions.
As such, no decoded graph contains connection types, i.e., triplets $(n,e,n)$ not present in the baseline embedding, even
though global properties such as topology and size differ significantly.
The consistency of global properties, however, varies between the models.
Note that for TREES samples, scaling the embedding introduces small cycles and hubs,
indicating that the embedding contains noisy dimensions that correspond to local structures
not present in the training dataset.
The effect may stem from the fact that the TREES dataset is the least constrained as colors
are assigned fully at random.
For the COLORING and PUBCHEM variants however, there are global constraints.
The graphs in COLORING have the properties of being planar, and in addition, no two neighbouring
nodes may share a color.
The graphs in PUBCHEM follow the laws of chemistry.
Accordingly, the model trained on the COLORING~15 variant decodes into graphs where no two adjacent nodes share a
color and which tend to be planar even when decoding into sizes not present in the dataset.
Similarly, the model trained on PUBCHEM~32 tends to repeat patterns present in the baseline as the embedding
is scaled.
The analysis in this section primarily aimed to inspect how modifications in the latent space
manifest in the graph space.
Models trained within our framework learn high-level topological properties of graphs such that latent-space
modifications cause changes in graph topology while preserving the distributional properties of the decoded graphs.
Recall that we train our model with two heads, one for the subgraph relationship and the other to predict
the set of possible action types.
As the model is trained to recognize a part-whole relationship, it is not surprising that the latent space
captures, at least approximately, topological and distributional properties of the underlying graph dataset.
An interesting question for future work is the extent to which GRAIL can be used as a generative
model for graphs.
However, this would likely require a more controlled optimization of the embedding space, for instance
through variational or contrastive objectives that impose additional structure on the latent representation.

\newcommand{\annotatedimage}[2]{%
    \begin{center}
    \begin{tikzpicture}
    \node[anchor=south west, inner sep=0] (img)
    {\includegraphics[width=0.95\linewidth]{#1}};
    \begin{scope}[x={(img.south east)}, y={(img.north west)}]
    \node at (0.05,1.25) {$\mathbf{y}$};
    \node at (0.14,1.25) {$\mathbf{\hat{y}}$};
    \node at (0.23,1.25) {$\mathbf{0}$};
    \node at (0.59,1.25) {$\mathbf{1z}$};
    \node at (0.96,1.25) {$\mathbf{#2z}$};
    \end{scope}
    \end{tikzpicture}
    \end{center}
}

\begin{figure}[htb!]
    \parbox{\linewidth}{
        \annotatedimage{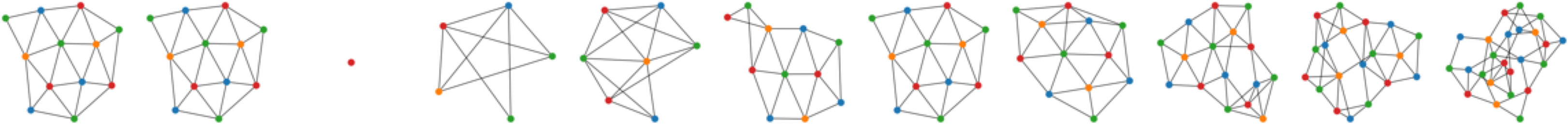}{2}
        \annotatedimage{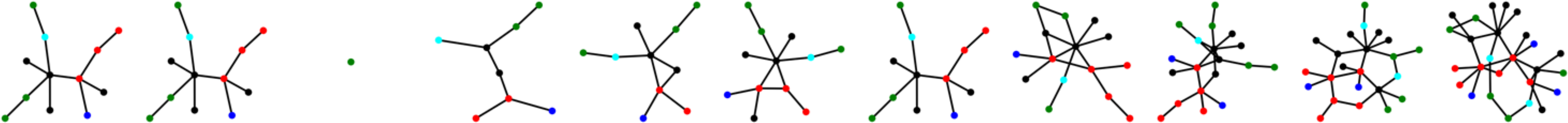}{3}
        \annotatedimage{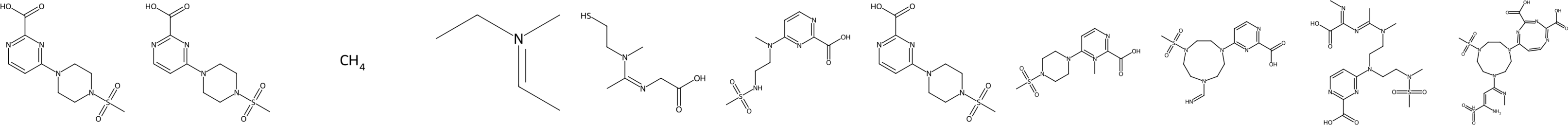}{2}
    }
    \caption{
        Graph-level interpolations between the origin $\mathbf{0}$ and $\mathbf{cz}$
        for a graph-embedding $\mathbf{z}$ and a constant $c$. The two leftmost graphs, $\mathbf{y}$
        and $\mathbf{\hat{y}}$, are the true and the decoded graphs, respectively.
        The decoded graph at $\mathbf{\hat{y}}=\mathbf{1z}$ is shown as a reference in the center of the
        interpolated graphs. Note that $\mathbf{y}=\mathbf{\hat{y}}$ only reliably holds whenever the model achieves
        high training accuracy.
    }
    \label{fig:interpolations}
\end{figure}

    \section{Limitations}\label{sec:limitations}
In this section, we address limitations of the current design and derive
further steps that are necessary to scale GRAIL to even larger graph instances.

\textbf{Forward vs. backward labeling.}
To produce training data for our models, we adopted a \emph{forward} view of starting
from an empty graph and completing a target.
To decide which successor states are valid partial results for a particular target, we introduced
subgraph matching as an expert policy, which works well for the graph classes and sizes we
studied in this work.
Even though algorithms for subgraph isomorphism have worst-case exponential time-complexity,
they tend to have a low average-case runtime when graphs have low symmetry, especially for labeled graphs.
However, scaling this approach to much larger sizes will likely increase the cost of the expert policy,
which may threaten the scalability of the approach.
While generation is inherently \emph{forward}, we can also adopt a \emph{backward} view for
labeling.
However, our framework does not prescribe any particular expert policy like checking for subgraph
isomorphism.
For each subgraph $G_{t}$, we can continue to enumerate the successors, but instead of running a subgraph check on each,
we use $G_{t}$ as a softmax target, making the training self-supervised.
A self-supervised setup would make the training similar to the next token prediction objective
in large language models, a training paradigm which is known to scale well with system size.

\textbf{Optimizing inference.}
To keep the search space tractable in our experiments, we presented filtering techniques,
one for filtering the types of expansions and one to avoid repeated expansions of non-subgraph
branches.
In our setup, we expand every remaining successor state.
However, for large graphs expanding all successors may be wasteful and one may choose to only expand one successor
at a time.
Furthermore, one may employ advanced expansion techniques like greedy expansions with backtracking,
similar to how subgraph matching algorithms match a target.
Another possibility to optimize inference latency is to expand graphs with motifs rather than
single elements.
While we presented the expansion of successor states at the most general level, our framework
naturally supports the incorporation of domain-specific knowledge.

\textbf{Summary.}
In this work, we studied the supervised graph prediction problem.
We first analyzed how recently proposed methods learn to predict
graphs as their outputs.
Through our analysis, we showed that while existing
models for supervised graph prediction can learn to predict graphs
in a supervised fashion, addressing graph isomorphism via alignment
methods may come with practical challenges, and as such, it may be worthwhile
to explore methods that do not rely on an explicit order.
We then presented a framework, \textbf{GRA}ph \textbf{I}mitation \textbf{L}earning (GRAIL),
for supervised graph prediction.
Our framework addresses the representational issues of contemporary
graph prediction models by treating graphs in an permutation invariant manner,
and the resulting design achieves encouraging results on a wide range of graph
prediction benchmarks.

    \bibliographystyle{plainnat}
    \bibliography{main}

    \appendix
    \section{Experimental setup}\label{sec:experimental-setup}
We primarily follow the experimental setup of~\cite{grale2025}.
However, for the synthetic TREES and COLORING datasets we do not use a fixed small dataset but
instead generate data on-the-fly.
The synthetic TREES dataset contains small colored trees, while COLORING~15 and COLORING~20
contain synthetic colored planar graphs satisfying the 4-color theorem.
The QM9~\cite{ramakrishnan2014quantum} dataset contains small molecules up to nine heavy atoms (nodes),
while the PUBCHEM~32 and PUBCHEM~40 contain larger molecules with up to 32 and 40 heavy atoms (nodes), respectively,
which are slices of the PubChem~\cite{kim2016pubchem} database.
However, PUBCHEM~32 is skewed towards medium-sized graphs and only a small fraction of samples
is within the upper size range.
This motivated us to assess the performance of our model on larger molecules only and to
create the PUBCHEM~40 slice which contains only samples with a minimum of 30 and a maximum of 40 nodes.

For fingerprint generation of molecules, we use Morgan Fingerprints with a radius of 2 and a size of 2048.
For the synthetic datasets, we implement the algorithm in listing~\ref{lst:lstlisting} as there is no general fingerprint algorithm for synthetic graphs.

\begin{lstlisting}[language=Python,label={lst:lstlisting}]
def get_graph_fingerprint(G: nx.Graph, size: int = 2048, radius=2) -> np.ndarray:
    bitvector = np.zeros(size, dtype=np.uint8)

    node_hashes = {}
    for n in G.nodes():
        node_type = str(G.nodes[n].get("type", "unknown"))
        label = f"type:{node_type}"
        node_hashes[n] = int(hashlib.md5(label.encode()).hexdigest(), 16)

        bitvector[node_hashes[n] % size] = 1

    for r in range(1, radius + 1):
        new_hashes = {}
        for n in G.nodes():
            neighbors_info = []
            for nbr in G.neighbors(n):
                edge_type = str(G[n][nbr].get("type", "default"))
                neighbors_info.append((edge_type, node_hashes[nbr]))

            neighbors_info.sort()

            combined = f"{node_hashes[n]}_" + "_".join(
                [f"{e}:{h}" for e, h in neighbors_info]
            )
            new_hash = int(hashlib.md5(combined.encode()).hexdigest(), 16)
            new_hashes[n] = new_hash

            bitvector[new_hash % size] = 1

        node_hashes = new_hashes

    return bitvector
\end{lstlisting}

\section{Hyperparameters}\label{sec:appendix-hyperparameters}

Here we report the hyperparameters for our experiments.
Hyperparameters are the same across all experiments, except that due to different computational requirements
in our time-bound experiments, some hyperparameters differ.
We summarize those below Table~\ref{tab:hparams}.

\begin{table}[h]
    \centering
    \begin{minipage}[t]{0.32\textwidth}
        \centering
        \begin{tabular}{|c|c|}
            \hline
            \textbf{Parameter} & \textbf{Value} \\
            \hline
            Network            & GINE           \\
            \hline
            \# Layers          & 5              \\
            \hline
            GNN dim.\           & 512            \\
            \hline
            MLP dim.\           & 512            \\
            \hline
        \end{tabular}
    \end{minipage}
    \hfill
    \begin{minipage}[t]{0.32\textwidth}
        \centering
        \begin{tabular}{|c|c|}
            \hline
            \textbf{Parameter} & \textbf{Value} \\
            \hline
            Network            & MLP            \\
            \hline
            MLP dim.\           & 256            \\
            \hline
            \# Layers          & 2              \\
            \hline

        \end{tabular}
    \end{minipage}
    \hfill
    \begin{minipage}[t]{0.32\textwidth}
        \centering
        \begin{tabular}{|c|c|}
            \hline
            \textbf{Parameter} & \textbf{Value} \\
            \hline
            Optimizer          & LAMB           \\
            \hline
            Loss               & Focal loss     \\
            \hline
            Focal $\gamma$     & $3.0$          \\
            \hline
            Focal $\alpha$     & $-1$           \\
            \hline
            LR start           & 1e-4           \\
            \hline
            LR max             & 1e-3           \\
            \hline
            LR end             & 1e-4           \\
            \hline
            LR warmup steps    & 1e9            \\
            \hline
            LR total steps     & 1e10           \\
            \hline
            Batch size         & 8k             \\
            \hline
        \end{tabular}
    \end{minipage}
    \caption{Hyperparameters for graph encoders (left), fingerprint encoders (center) and training (right).}
    \label{tab:hparams}
\end{table}

We use a pre-trained ResNet50 as an image encoder and shorten the LR schedule to peak at 100M samples and use a batch size
of 2048.
For the MLP decoder, we use [2048,2048,1024,1024] units for the subgraph prediction head and [1024,1024] for the action filter head
with a dropout value of 0.1.
In all non-image tasks, we use LayerNorm across all modules.

\section{Training statistics}\label{sec:appendix-training}

\begin{figure}[htb]
    \centering
    \begin{subfigure}{0.48\textwidth}
        \centering
        \includegraphics[width=\linewidth]{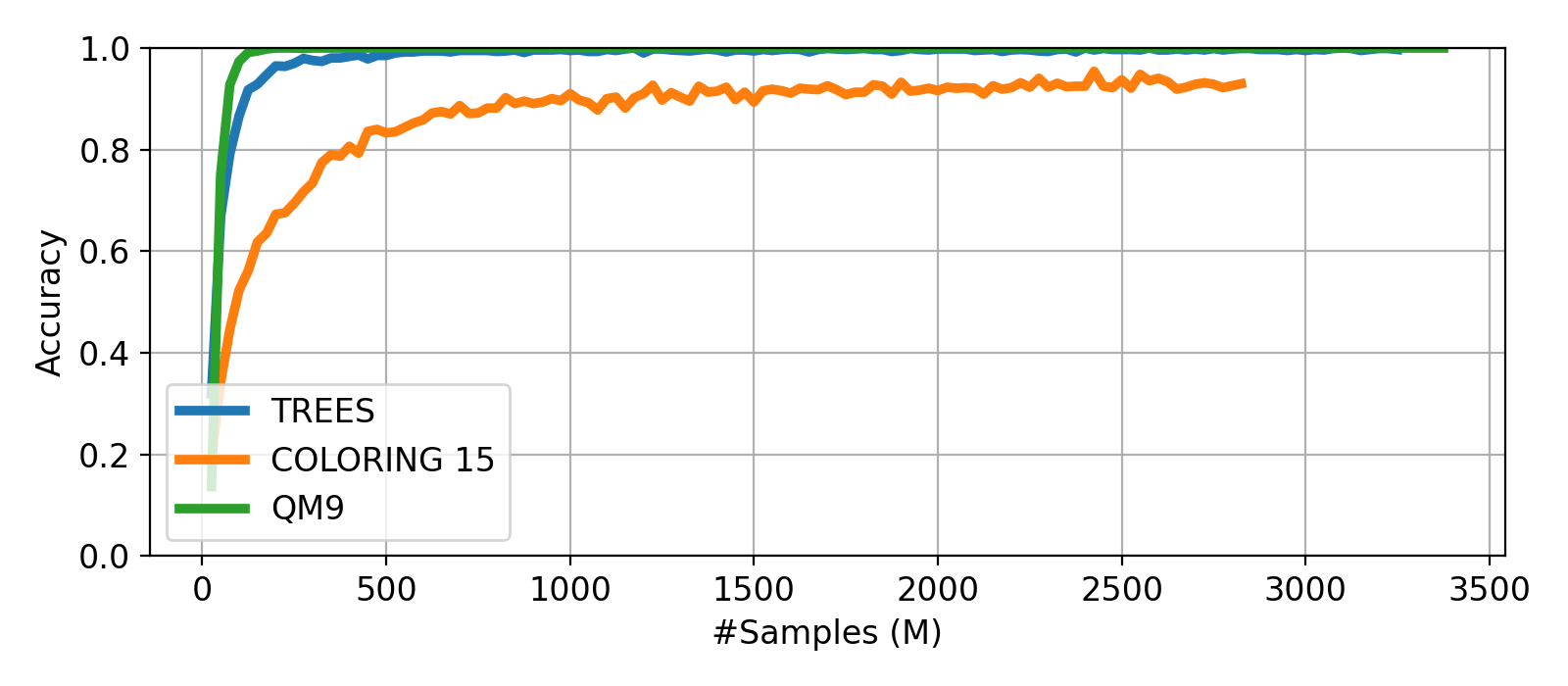}
        \caption{Training on graphs.}
    \end{subfigure}
    \hfill
    \begin{subfigure}{0.48\textwidth}
        \centering
        \includegraphics[width=\linewidth]{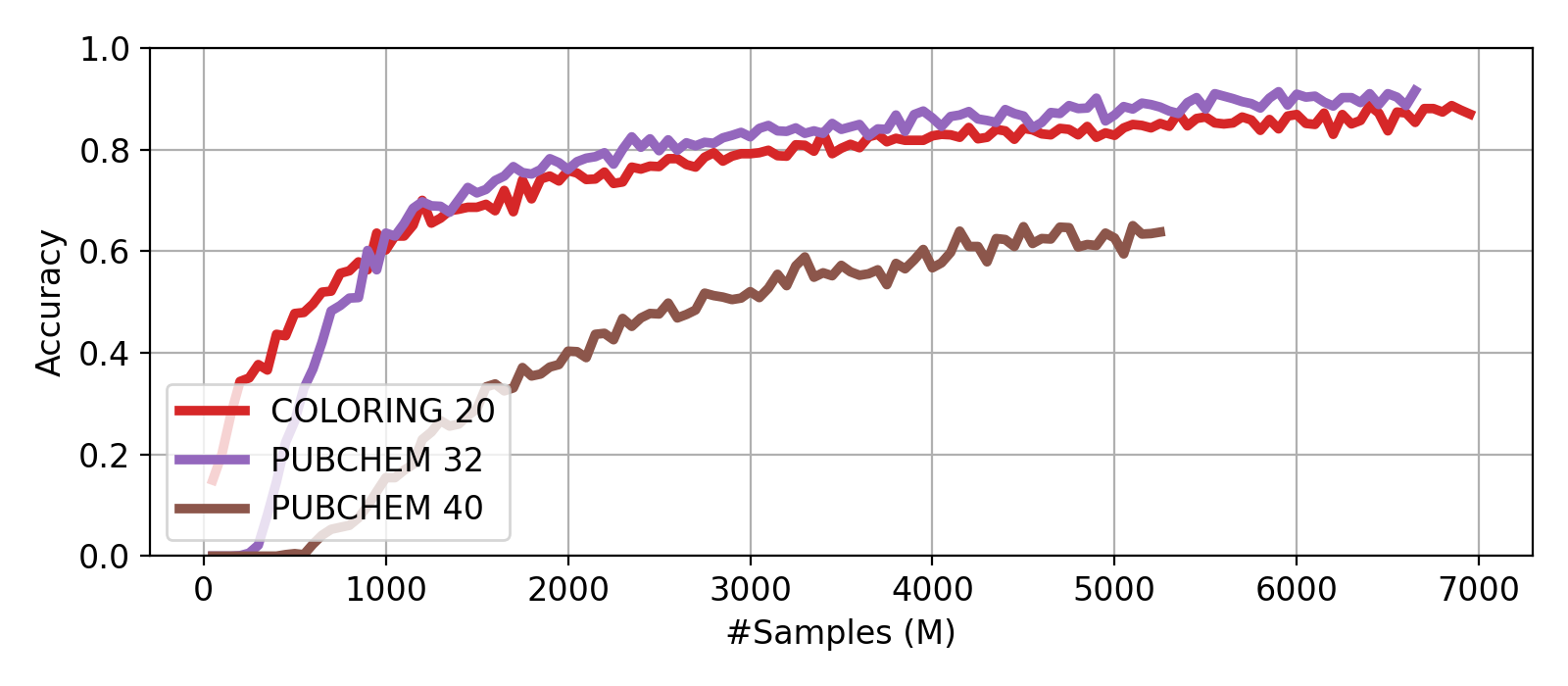}
        \caption{Training on graphs.}
    \end{subfigure}

    \vspace{0.5em}

    \begin{subfigure}{0.48\textwidth}
        \centering
        \includegraphics[width=\linewidth]{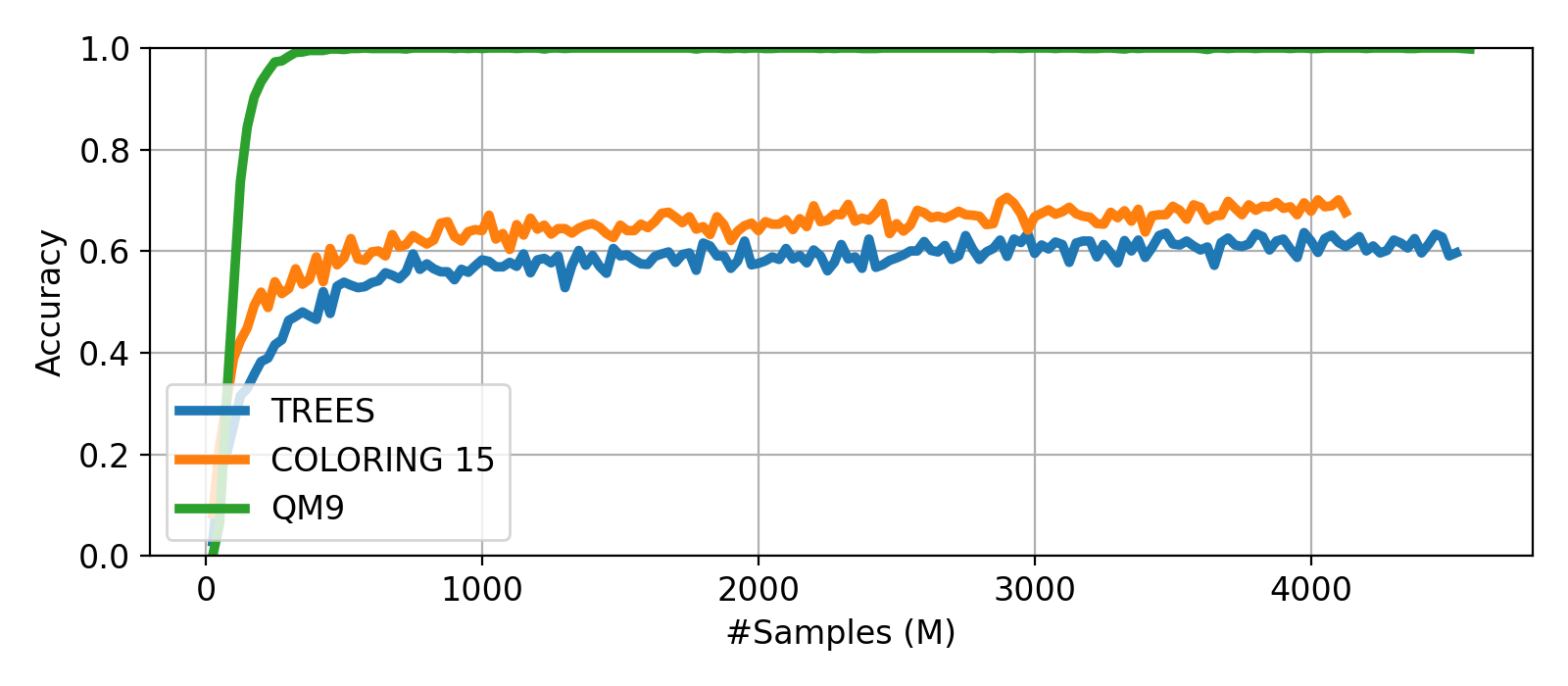}
        \caption{Training on fingerprints.}
    \end{subfigure}
    \hfill
    \begin{subfigure}{0.48\textwidth}
        \centering
        \includegraphics[width=\linewidth]{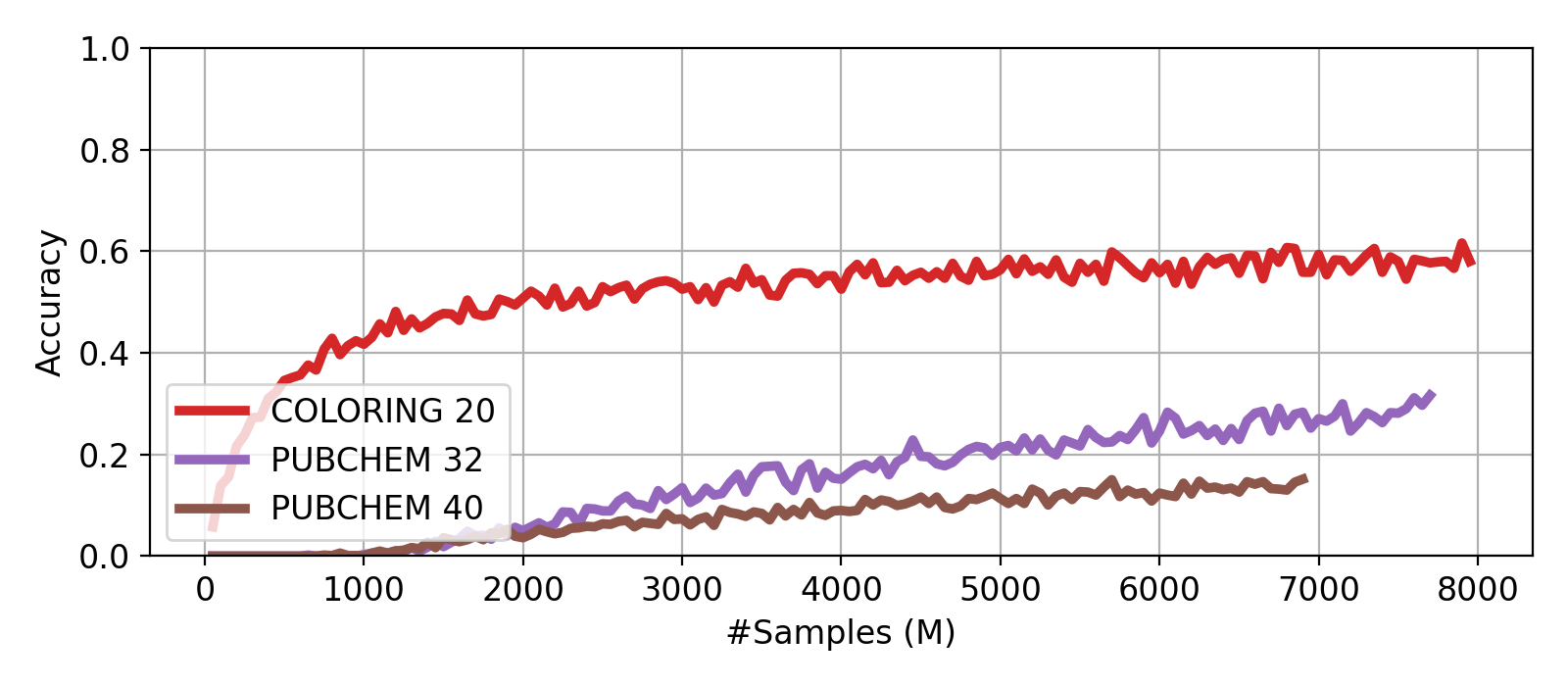}
        \caption{Training on fingerprints.}
    \end{subfigure}

    \begin{subfigure}{0.48\textwidth}
        \centering
        \includegraphics[width=\linewidth]{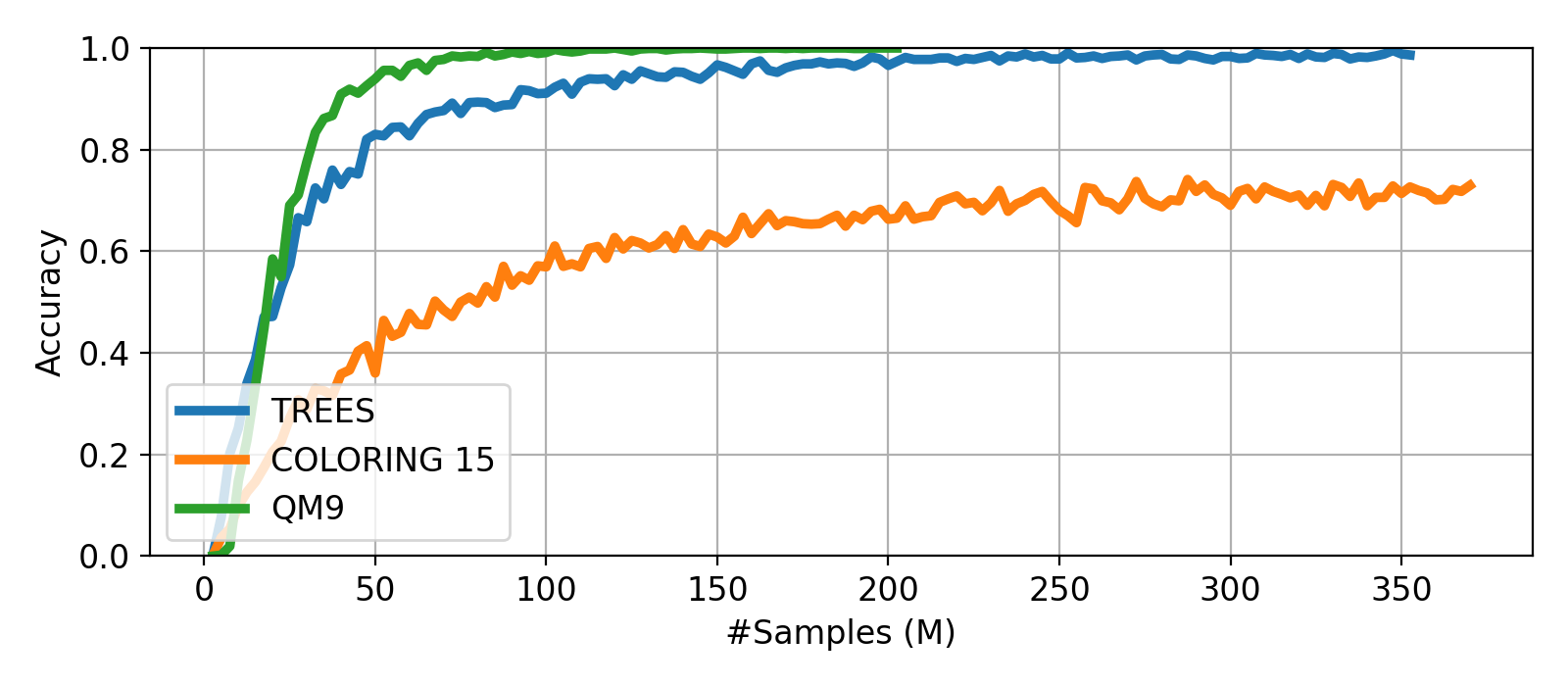}
        \caption{Training on images.}
    \end{subfigure}
    \hfill
    \begin{subfigure}{0.48\textwidth}
        \centering
        \includegraphics[width=\linewidth]{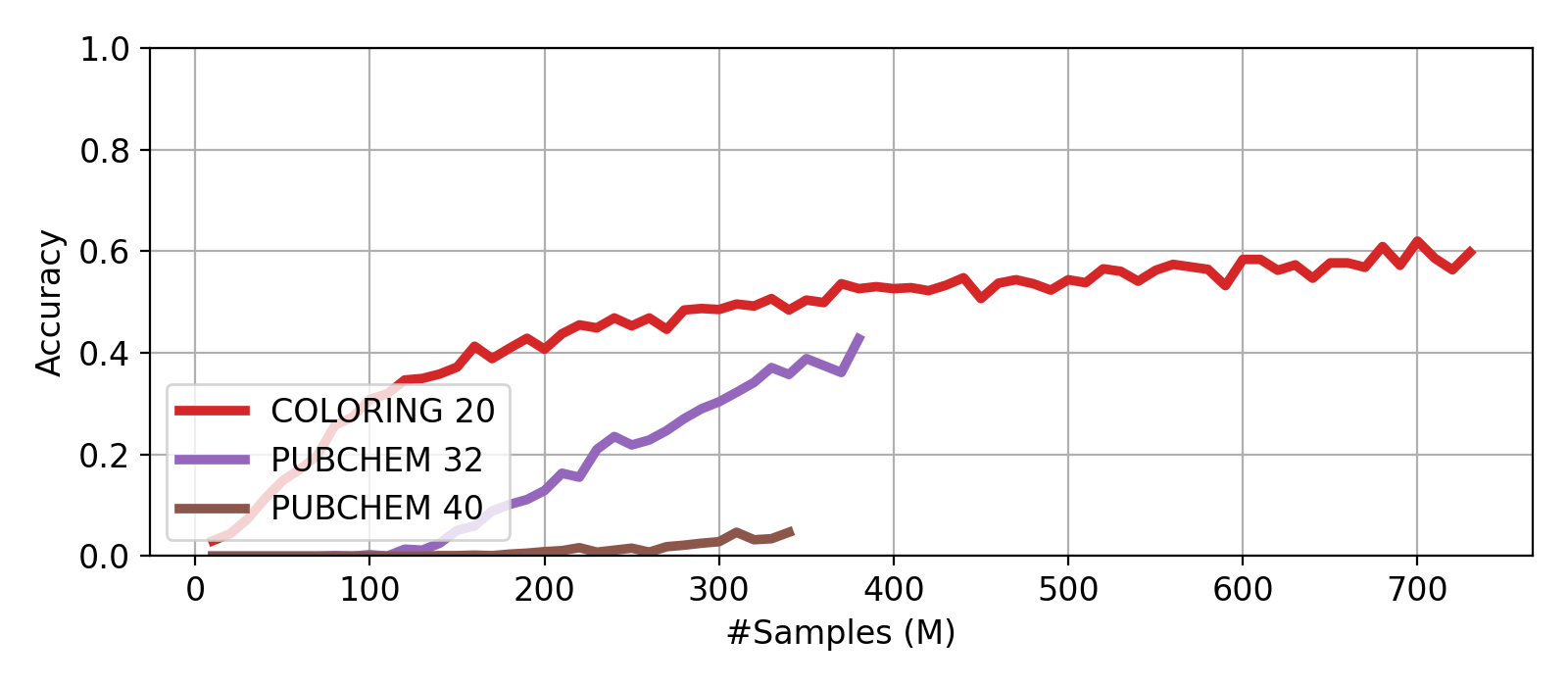}
        \caption{Training on images.}
    \end{subfigure}

    \caption{Training progress of all models trained on graphs, fingerprints and images.}
    \label{fig:appendix-training}
\end{figure}

\paragraph{Training process.}
Figure~\ref{fig:appendix-training} shows the training curves for all experiments.
The experiments have been conducted in a time-bound setting with a fixed training
time of 24h.
We trained the models on the smaller datasets, TREES, COLORING 15, and QM9 on four A100 GPUs,
while we trained the models on the larger datasets, COLORING 20, PUBCHEM 32, and PUBCHEM 40,
on 16 A100 GPUs.
Training models under our framework is generally stable.
For small datasets such as QM9, we can achieve perfect results within a very short timeframe.
For larger and more complex datasets like large PUBCHEM slices, training becomes slower as
the number of samples from which the model can learn effectively shrinks over time.
Therefore, in large spaces it may become necessary to select samples for training less uniformly.
Note that for image models, the number of processed samples is much lower due to the computational cost of encoding images.

\begin{figure}[htb]
    \centering
    \includegraphics[width=\linewidth]{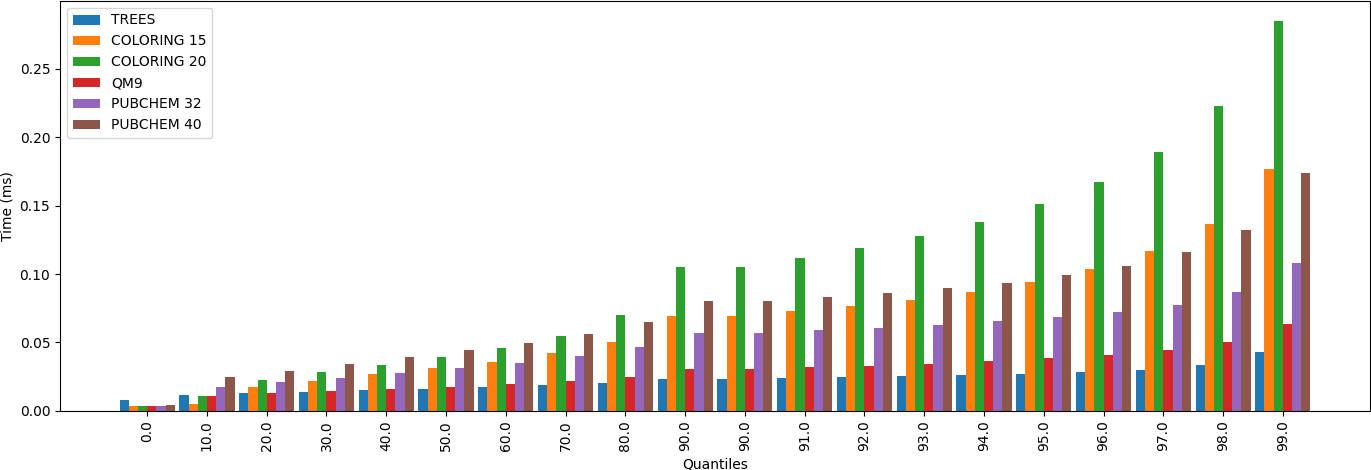}
    \caption{Time for expert policy calculation up to the 99th quantile.}
    \label{fig:appendix_quantiles_99}
\end{figure}

\begin{figure}[htb]
    \centering
    \includegraphics[width=\linewidth]{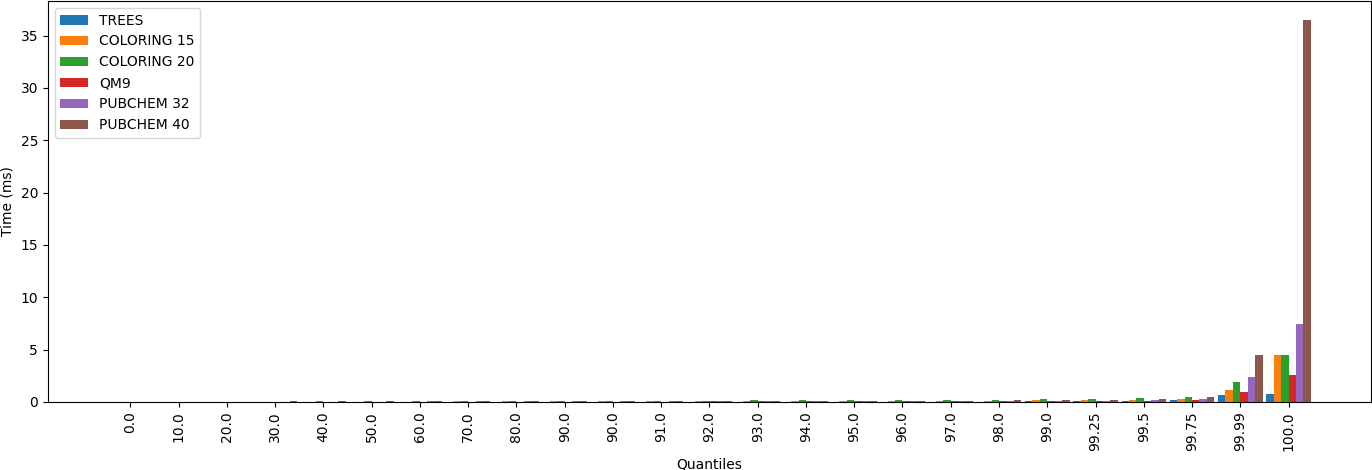}
    \caption{Time for expert policy calculation up to the 100th quantile.}
    \label{fig:appendix_quantiles_100}
\end{figure}

\paragraph{Expert policy.}
In Figures~\ref{fig:appendix_quantiles_99} and~\ref{fig:appendix_quantiles_100} we show for
all datasets the distribution of calculation times for the expert policy.
Note that the quantiles are not linear and focus on the upper quantiles.
As such, generating training data is, at least in these classes of graphs, not data-bound
and can be trivially parallelized to generate huge amounts of data for training.
Training efficiency is mostly driven by the long tail of rare type distributions.


\paragraph{Search space.}

\begin{figure}[htb]
    \centering
    \includegraphics[width=\linewidth]{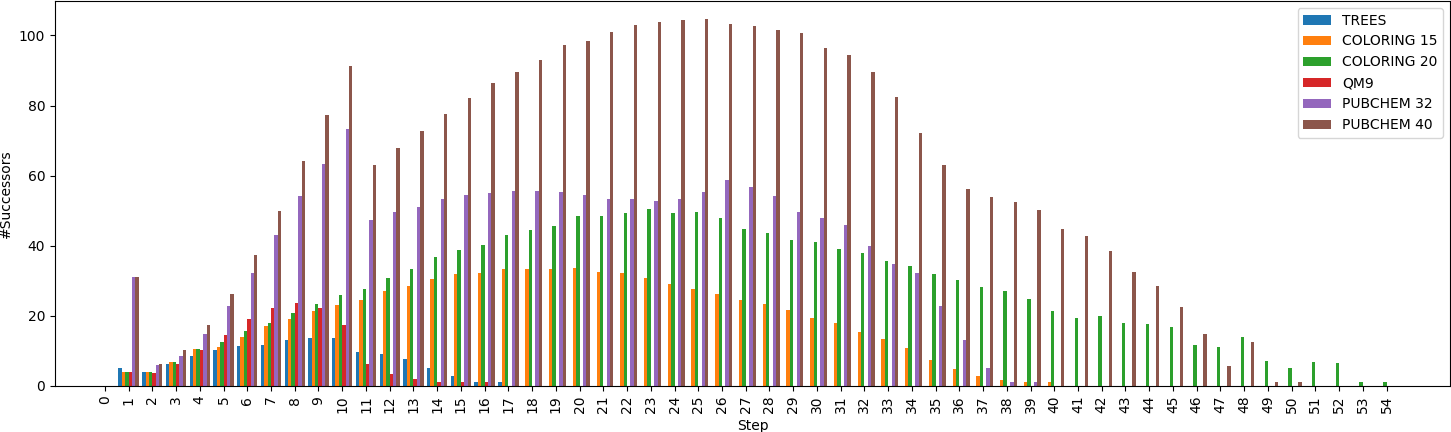}
    \caption{Number of successor states per step for each dataset.}
    \label{fig:appendix_successors}
\end{figure}

In Figure~\ref{fig:appendix_successors}, we show for all datasets the distribution of
the number of successor states based on the step.
In general, the curves are bell-shaped with the peak roughly in the middle of the trajectory.
This behavior stems from how our model filters the successor states.
Once a particular type of connection has been satisfied, they are not expanded further.
Expansions which the model predicted as invalid transitions in earlier states are not enumerated repeatedly.
Together, this creates a strong filtering effect and bounds the number of successor states
which would, without any mitigations, explode for long trajectories.

\section{Pseudocode for inference in GRAIL}\label{sec:pseudocode}
\begin{algorithm}[t]
\caption{GRAIL inference}
\label{alg:grail}
\begin{center}
\begin{minipage}{0.95\linewidth}
\textbf{Input:} $x_T$, encoders $f_T, f_Q$, policy $\pi$, filter $f_E$ \\[2pt]
$\mathbf{z_T} \gets f_T(x_T)$ \\
$G_0 \gets \emptyset$ \\[4pt]

\textbf{for} $t = 0,1,2,\dots$ \textbf{do} \\
\hspace*{1em} $\mathbf{z_Q} \gets f_Q(G_t)$ \\
\hspace*{1em} $\mathbf{m} \gets f_E(\mathbf{z_Q}, \mathbf{z_T})$ \\
\hspace*{1em} $\mathbb{M}_t \gets \emptyset$ \\

\hspace*{1em} \textbf{for each} $(n_u,e,n_v) \in \mathcal{A}$ with $\mathbf{m}_{(n_u,e,n_v)} = 1$ \textbf{do} \\
\hspace*{2em} \textbf{for each} valid application of $(n_u,e,n_v)$ to $G_t$ \textbf{do} \\
\hspace*{3em} $\mathbb{M}_t \gets \mathbb{M}_t \cup \{G'\}$ \\
\hspace*{2em} \textbf{end for} \\
\hspace*{1em} \textbf{end for} \\

\hspace*{1em} $\mathbb{M}_t \gets \mathbb{M}_t \cup \{G_t\}$ \\
\hspace*{1em} $(term, G_{t+1}) \gets \arg\max\limits_{G' \in \mathbb{C}_t} \ \pi\big(f_Q(G'), \mathbf{z_T}, term\big)$ \\
\hspace*{1em} $G_{t} \leftarrow G_{t+1}$ \\

\hspace*{1em} \textbf{if} $term = 1$ \textbf{then break} \\
\textbf{end for} \\[4pt]

\textbf{Return:} $G_t$
\end{minipage}
\end{center}
\end{algorithm}

%    \clearpage
%    \input{checklist.tex}

\end{document}